# Hybrid 2-stage Imperialist Competitive Algorithm with Ant Colony Optimization for Solving Multi-Depot Vehicle Routing Problem

Ivars Dzalbs, Tatiana Kalganova[1]

1 Brunel University London, Kingston Lane, Uxbridge, UB8 2PX, UK
Tatiana.Kalganova@brunel.ac.uk

**Abstract.** The Multi-Depot Vehicle Routing Problem (MDVRP) is a real-world model of the simplistic Vehicle Routing Problem (VRP) that considers how to satisfy multiple customer demands from numerous depots. This paper introduces a hybrid 2-stage approach based on two population-based algorithms – Ant Colony Optimization (ACO) that mimics ant behaviour in nature and the Imperialist Competitive Algorithm (ICA) that is based on geopolitical relationships between countries. In the proposed hybrid algorithm, ICA is responsible for customer assignment to the depots while ACO is routing and sequencing the customers. The algorithm is compared to non-hybrid ACO and ICA as well as four other state-of-the-art methods across 23 common Cordreau's benchmark instances. Results show clear improvement over simple ACO and ICA and demonstrate very competitive results when compared to other rival algorithms.

## 1. Introduction

The Vehicle Routing Problem (VRP), first described by [1] in 1959, is an extension of the Traveling Salesman Problem (TSP) [2]. Compared to TSP, where an agent has only to visit all cities once, VRP introduces demands for each customer or stop. Demands need to be satisfied by routing vehicles such that they start and finish their paths at the same depot. Many real-life problems can be modelled as a form of VRP, for example, picking up and delivering mail, packages or any other goods or services. Due to the wide range of practical applications, many variations of VRP have since been explored. For instance, capacitated VRP introduces capacity constraints on the vehicles; VRP with Time Windows (VRPTW) requires delivery to happen within a specific time window; VRP with maximum vehicle distance constraints (DVRP) and many others [3].

A common VRP derivation is the Multi-Depot Vehicle Routing Problem (MDVRP). MDVRP is an extension of classical VRP by the introduction of multiple depots. Vehicles in the MDVRP are subject to capacity constraints (how much cargo can be carried onboard) and the maximum duration for the route before the vehicle needs to return to the original depot. The MDVRP resembles a lot of everyday transportation, logistics and distribution problems and therefore, has been a common research area [4]. Furthermore, the MDVRP is also an NP-hard combinatorial optimization problem, thus optimal solutions are hard to find [5]. Although exact algorithms for solving these class of problems exist, they are limited to small problem instances [6]. To solve large instances of the MDVRP, wide range of metaheuristics and population-based algorithms have been used [4]. This paper explores the use of a combination of two stand-alone metaheuristics to tackle the MDVRP.

The main contribution of this paper is the novel hybrid 2-stage Imperialist Competitive Algorithm (ICA) with Ant Colony Optimization (ACO) called ACO-ICA as an improvement of its algorithm counterparts for solving the MDVRP. In the proposed approach, ICA is responsible for customer clustering to their depots while ACO is focused on routing and sequencing the customers. Furthermore, computation results and comparisons for 23 commonly used MDVRP instances are provided for both ACO and ICA, with ICA applied to MDVRP for the first time. Moreover, the computation results of ACO-ICA are also compared to the most recent state-of-the-art metaheuristics found in the literature.

## 1.1. Multi-Depot Vehicle Routing Problem formulation

The main aim of the MDVRP is to route a fleet of vehicles from multiple depots to multiple customers requiring goods or services. Vehicles delivering the products start from a depot and are required to terminate at the same depot while complying with constraints such as maximum vehicle capacity and maximum vehicle duration. Figure 1 shows an example of a simple MDVRP solution with ten customers (as circles) and two depots (as rectangles). Although there exist multi-objective approaches for solving MDVRP [7], the most common goal is to minimize the total cost.

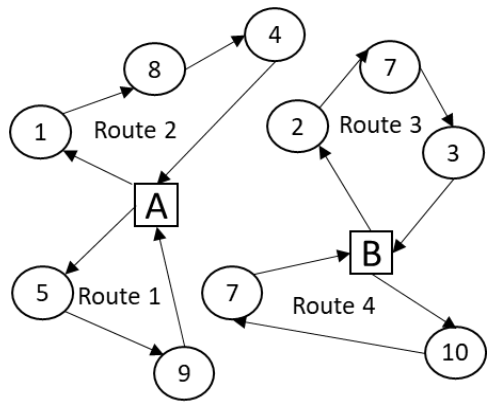

Figure 1. Example of an MDVRP with ten customers (as circles) and two depots (A and B as rectangles)

The MDVRP can be formalized in a mathematical model based on [8] and [9]. Given direct graph $G = (S, E)$ where $S = C \cup U$ is a set of customers $C = \{C_1, C_2, \ldots, C_N\}$ and depots $D = \{D_1, D_2, \ldots, D_M\}$ and $E$ is a set of edges between all the nodes in the graph. In a fully connected graph, every edge $E_{ij}$ between nodes $S_i$ and $S_j$ $(i \neq j)$ has associated positive cost $c_{ij}$ - distance or time, for example. Each customer has a positive demand $d_i$ $(i \in C)$. Furthermore, there is also a fleet of $K$ identical vehicles available at each depot $D_k \in D$ (that are not allowed to exceed capacity $Q_{max}$ and duration $R_{max}$). The goal is to minimize the total cost across all vehicles (1).

$$\min \sum_{i \in S} \sum_{j \in S} c_{ij} x_{ij} \qquad (1)$$

where $x_{ij}$ equals 1 if $i$ comes after $j$ in the customer sequence on any route of all vehicles and 0 otherwise. The problem is subject to the following constraints:

- each vehicle route starts and ends at the same depot;
- the total demand on each route does not exceed vehicle capacity $Q_{max}$;
- the maximum route duration $R_{max}$ is not exceeded;
- each customer is served by exactly one vehicle.

## 2. State-of-the-art algorithms for solving the MDVRP

Since the first formulation in [1], many exact and heuristic algorithms have been explored for the vehicle routing problems. Most notably, [10] proposed a heuristic approach based on the cost savings algorithm that has since been used in some form in many other algorithms [11]. Another popular heuristics approach was introduced in [12] that allowed problems divided into sub-problems based on vehicles and then solved separately, combining results into single solution afterwards. Although heuristic approaches such as integer programming [13] and variable neighbourhood search [14] have the potential to find optimal solutions every time, they generally do not scale well with the problem size and are limited to smaller MDVRP instances or are very time-consuming [6].

Meta-heuristic algorithms offer a stochastic approach for solving highly complex combinatorial problems with near-optimal or optimal solutions. They have been a growing interest in many areas [15], and MDVRP is no exception. A recent survey of metaheuristic algorithms [4] suggests that the two of the most common algorithms used for solving MDVRP are Ant Colony Optimization (ACO) and Genetic Algorithm (GA). However, other algorithms like Particle Swarm Optimization (PSO) [16] and Ant Lion Optimization (ALO) [17] have also been successfully applied. GA is a nature-inspired algorithm that is based on the natural selection process. A comprehensive summary of methods and approaches used for solving MDVRP with GA is presented in [3]. ACO is another popular approach for solving VRP class problems as it mimics ants travelling and searching for food while creating paths for other ants to follow. Many implementations of ACO for MDVRP exist in the literature; the most recent work includes [18] who applied the ACO algorithm for fresh seafood delivery routing problem.

More recently hybridized algorithms have emerged that take one or more methodologies and combines the strengths of each. Most commonly meta-heuristic algorithm is combined with a local search such as 2-opt [17] or simulated annealing ([9], [19]). Similarly, [20] introduced a mutation operator from GA in ACO to improve the solution quality in MDVRP.

## 3. Hybrid 2-stage ACO-ICA

The Hybrid 2-stage Imperialist Competitive Algorithm with Ant Colony Optimization (ACO-ICA) combines an ICA based algorithm previously proposed by authors in [21] for customer assignment to depots. At the same time, ACO (also already used by authors in [22]) is dedicated for customer sequencing and routing in the MDVRP.

### 3.1. Ant Colony Optimization

The Ant Colony Optimization (ACO) algorithm was first introduced by Marco Dorigo in 1992 and since been used for many routing problems [23]. The algorithm is inspired by the movement of ants searching for food; each ant leaves a scent called pheromone for other ants to follow. An algorithm can be divided into two stages – the construction of the solution and pheromone feedback to other ants. The search for food starts at near-random, but as more pheromone get deposited on good paths, other consecutive ants are more likely to follow them while improving the solution. The iterative process continues until the solution reaches the optimal or close to optimal solution.

In the MDVRP, the ant starts the tour from a depot and visits nodes (customers) and returns to the same depot. As there are multiple depots, the algorithm uses multiple ant colonies (one per each depot) to solve the problem. Once all customers have been visited, routes from all colonies are combined and evaluated as the final solution. For paths that improved cost, a pheromone is deposited on the edges of the graph to guide ants in the next iteration. Furthermore, an evaporation process takes place after each iteration to avoid getting stuck in local optima. During the creation of the solution, an ant can visit any of the customers that have not yet been visited.

## 3.2. Imperialist Competitive Algorithm

The Imperialist Competitive Algorithm (ICA) was first introduced in 2007 for solving continuous optimization problems [24] but has been extended to other discrete problems. The ICA is inspired by geopolitical relationships between countries where developed countries attempt to take over less developed countries – colonize them to extend their power [25]. The initial population of countries is divided into imperialists and colonies based on the country's strength (inverse of cost value). After the grouping, each colony inside empire moves closer to their imperialist (assimilation operator). Imperialists compete for the colonies and the imperialist competition gradually results in an increase in power of strong empires while decreasing the power of weak ones. Weak empires eventually lose all colonies and collapse, with the strongest remaining imperialist providing the final best solution [24].

Although no ICA implementation for solving the MDVRP with ICA exists in the literature, the algorithm was applied for VHR with time windows in [26] where the authors encoded the routes in the country based on the vehicle. Furthermore, authors in [27] used slightly different encoding, where not only the visited customer sequence but also the number of visited customers is embedded in the country representation. In contrast to other approaches, the ICA implementation in this paper encodes the depots as part of the country, not the customers. An example of the ICA encoding is shown in Figure 2.

## 3.3. Hybrid ACO-ICA 2-stage algorithm

There have been few attempts to hybridise both ACO and ICA with other algorithms. Compared to ICA, the ACO algorithm has been around for longer and hence many more hybrids exist and a survey in [28] provides an excellent summary. Although not as common, ICA hybrids have been explored before. For instance, the authors in [29] combined ICA with particle swarm optimization (PSO) for the management of reactive power resources. Furthermore, [30] solved a facility relocation problem by combining ICA with simulated annealing and local neighbourhood search. Combination of ACO and ICA has also been attempted before in [31] for truss structure design where ACO was used to improve on ICA solutions. Compared to the ACO and ICA hybrid in [31] where ACO is used as local search, the approach in this paper is separating the creation of the solution in two stages – first, ICA assigns the customers to depots and then ACO routes and finishes the final solution.

The proposed ACO-ICA 2-stage hybrid algorithm uses the ICAwICA algorithm previously developed by authors in [21] for the customer assignment process. The search starts by empire initialization, where random solutions are created, and countries are split into imperialist and colonies. The search then follows an iterative process of colony assimilation and imperialist competition as described in [21] and the flowchart in Figure 4. Each new country created by ICA follows the 2-stage process for creating a feasible solution:

**Stage 1: Customer assignment to depots**

First, customer-depot relationships are encoded as the country. Each country is represented as a vector of the size of the number of customers where each customer is assigned a depot index. An example of new country creation via assimilation is shown in Figure 2, where the initial colony has encoded the following grouping: Customer 2 and 8 will be routed from Depot 1; Customers 1, 3 and 6 will be routed from Depot 2; Customers 5,7,9 and 10 will be routed from Depot 3 and finally, Customer 4 will be routed from Depot 4. Each time new country is created as part of ICA assimilation process, capacity constraints are considered such that the total demand for all customers assigned to the depot does not exceed the maximum capacity available across all vehicles to the given depot.

![Figure 2 diagram showing Colony, Imperialist, and New country rows with depot indices for customers C1-C10]

*Figure 2. Customer assignment to depots using ICA assimilation in ACO-ICA 2-stage algorithm. Where C1-C10 are customer indices and the encoded integers are depot indices that are assigned to a given customer. With bold representing assimilated changes.*

Furthermore, the example in Figure 2 also shows an assimilation process for the colony and imperialist and considers ten customers that are grouped into four depots. Bold type represents assimilated changes. For example, Customer 2 (C2) demand was previously supplied by Depot 1 but now is supplied by Depot 4. Similarly, Customer 6 (C6) demand was previously supplied by Depot 2 but now is supplied by Depot 3.

**Stage 2: Customer routing**

Once customers have been assigned to the depots, an ant colony for each depot is created. Furthermore, persistent global pheromone matrix for the country is used to guide the colony search, where the pheromone matrix is represented as all connections $E$ between two edges in the graph $G$. The probability of visiting a customer that is not assigned to the depot is set to 0, therefore, no customer can be visited from more than one depot. The ants in each depot's ant colony build routes complying with the MDVRP constraints and update the local pheromone matrix. This process is repeated until the termination condition is reached. A flowchart of ACO customer routing in the MDVRP is shown in Figure 4.

Partial solutions for each depot are then combined and the final solution evaluated. The global pheromone matrix for the country is updated based on the best solution and global pheromone evaporation operator. The pseudo-code for new solution creation is shown in Figure 3.

```
Stage1: Create new country with ICA
Stage1: Split customers to depots while maintaining capacity constraints
for D = 1 to D_M
    Stage 2: route customers C allocated at depot D with ACO
end loop
Combine depot partial solutions and evaluate cost
if solution is better than the current country's best solution
    Keep best solution
endif
Update global pheromone matrix for ACO based on the best solution
Evaporate global pheromone
```

*Figure 3. Pseudo code for new solution creation of hybrid ACO-ICA 2-stage algorithm*

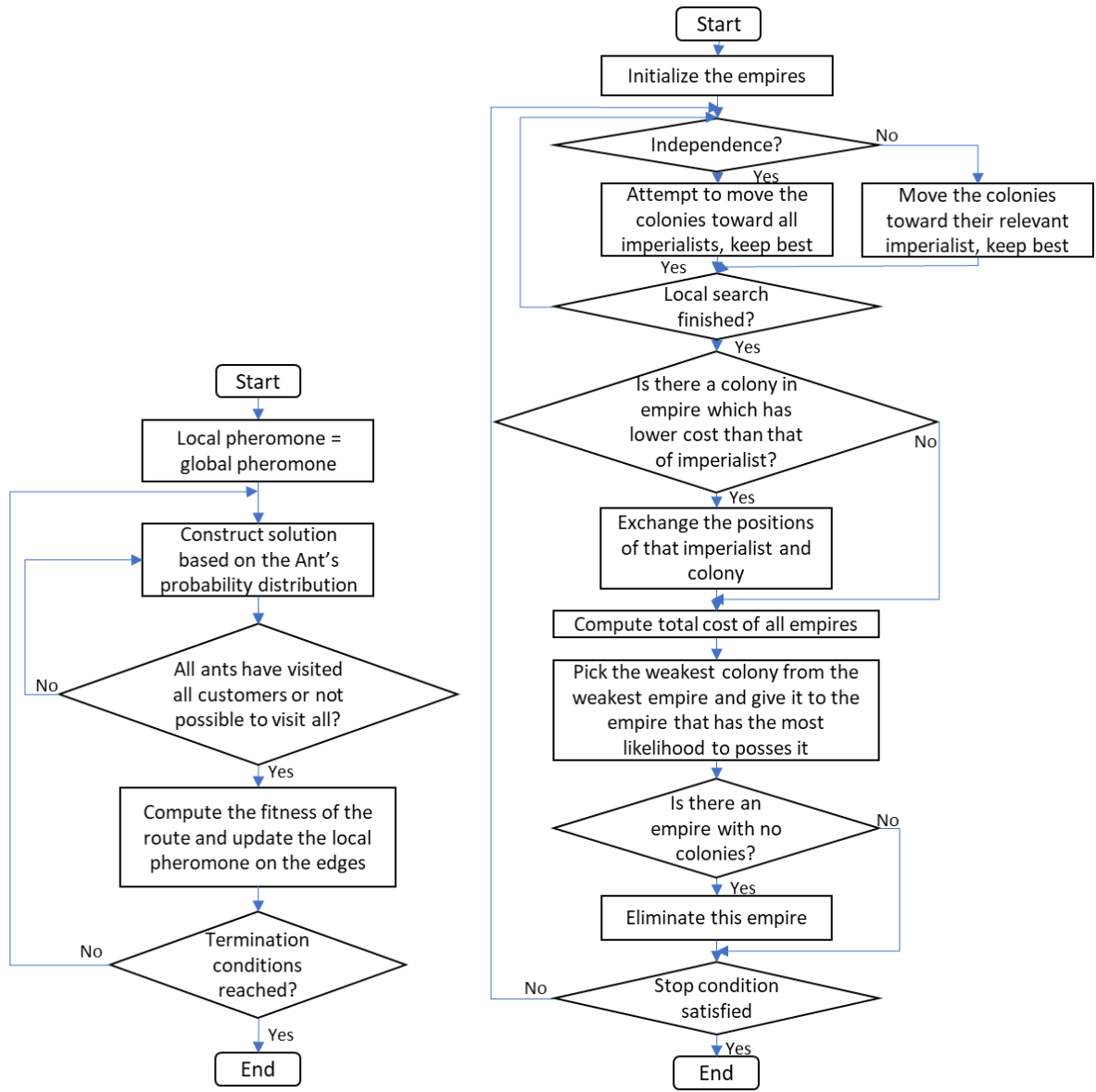

*Figure 4. Flowchart of Ant Colony Optimization for route creation in ACO-ICA 2-stage algorithm (on the left). And flowchart of ICAwICA [21] algorithm on the right.*

## 4. Experimental results

The proposed ACO-ICA hybrid algorithm was evaluated on well-known Cordeau's [32] MDVRP benchmark instances. The dataset with the best-known solutions (BKS) was obtained from [33]. Table 2 specifies the properties of each of the 23 instances. In the table, $N$ represent the number of customers, $M$ – the number of depots, $Q_{max}$ – maximum vehicle capacity and $R_{max}$ - maximum vehicle duration for given problem instance.

All three algorithms – ACO (based on [22]), ICA (based on [21]) and ACO-ICA were implemented in C++ using Visual Studio 2019 (v142) compiler. The computation was performed on a workstation with AMD Threadripper 2990WX processor (3.0 GHz, 64GB RAM), running Windows 10 Pro operating system. Algorithm hyperparameters were chosen with both empirical study and from the relevant literature ([9], [21], [31]) and are as summarized in Table 1. Furthermore, termination criteria set to the maximum number of iterations with no improvement (stagnation iterations, $n_i$) or one hour of elapsed time, whichever terminates first.

Table 1. Hyper-parameters used for experiments for ACO, ICAwICA and ACO-ICA

| ACO | |
|---|---|
| Number of ants, $n_a$ | 10 |
| Relative pheromone strength, α | 2 |
| Relative heuristic information strength, β | 1 |
| Pheromone evaporation rate, ρ | 0.1 |
| Pheromone update rate, σ | 0.1 |
| Exploitation to exploration ration, $q0$ | 0.5 |
| Stagnation iterations, $n_i$ | 20000 |
| **ICAwICA** | |
| Number of countries, $N_{population}$ | 4096 |
| Number of imperialists, $N_{imperialists}$ | 1638 |
| Local iterations, $N_{localIter}$ | 16 |
| Assimilation rate, ϒ | 0.05 |
| Average power of empire's colonies, $\xi$ | 0.05 |
| Independence rate, $independanceRate$ | 0.7 |
| Stagnation iterations, $n_i$ | 10 |
| **ACO-ICA** | |
| Number of countries, $N_{population}$ | 128 |
| Number of imperialists, $N_{imperialists}$ | 51 |
| Local iterations, $N_{localIter}$ | 1 |
| Assimilation rate, ϒ | 0.1 |
| Average power of empire's colonies, $\xi$ | 0.05 |
| Independence rate, $independanceRate$ | 0.8 |
| Number of ants, $n_a$ | 3 |
| Relative pheromone strength, α | 4 |
| Relative heuristic information strength, β | 1 |
| Pheromone evaporation rate, ρ | 0.1 |
| Pheromone update rate, σ | 0.1 |
| Exploitation to exploration ration, $q0$ | 0.8 |
| ACO iterations per country creation, $n_{aco}$ | 100 |
| Stagnation iterations, $n_i$ | 50 |

All three algorithms were run on the 23 Cordeau's MDVRP benchmark instances 10 times with different random seed and results summarized in Table 2, where BKS represents the best-known solution obtained from [33]. The best-obtained values, as well as average values across all 10 runs, were recorded alongside the average computational times (in minutes) for a single execution of the algorithm. With * representing values that were terminated due to the maximum allowed execution time of one hour. Finally, the average error percentage across all 23 instances is calculated using BKS as reference.

Results in Table 2 show that ACO algorithm was able to reach the best-known solution on four instances (p01, p12, p13 and p16), ICA reached the best-known solution on nine cases (p01, p02, p03, p06, p12, p13, p16, p17 and p19). The hybrid approach of ACO-ICA performing the best – improving on the best-known solution on one instance (p11) and finding the same solution in 16 cases (p01, p02, p03, p04, p06, p08, p09, p10, p12, p13, p14, p15, p16, p17, p19, p20). The average best error across all benchmark instances was 1.2%, 0.67% and 0.2% for ACO, ICA, ACO-ICA respectively. It is also worth noting that ACO took the least execution time, while ICA – the most, on nine problems reaching the maximum allowed time. The hybrid ACO-ICA approach on average was faster than the ICA, but slower than ACO, however, outperformed both non-hybrid algorithms in terms of average error.

Next, the hybrid ACO-ICA algorithm was compared to other state-of-the-art algorithms. Although there have been many algorithms applied to the MDVRP, the most recent approaches in literature were selected and are summarized in Table 3. A cooperative coevolutionary algorithm called CoES [34], Improved Ant Colony Optimization (IACO) [18], Tabu Search Heuristic (TSH) in [35], as well as hybrid Ant Colony with simulated annealing and local search algorithm called ACO+ [9] were selected for the comparison. The ACO-ICA algorithm was also compared to the best-known solutions in [33] and it is worth mentioning that these solutions are outdated as better results are reported in the literature. Nevertheless, the best-known solutions of [33] are included for reference.

Table 2. Computational results of Cordeau's MDVRP benchmark instances. Best and Average solution scores are derived across 10 independent runs, with the average time to reach a solution (in minutes). With bold representing the best scores. Time with * representing solutions that were terminated by reaching maximum allowed time (1h). Average error percentage calculated using BKS as reference.

| Instance | BKS [33] | Best ACO | Best ICA | Best ACO-ICA | Average ACO | Average ICA | Average ACO-ICA | Time (minutes) ACO | Time (minutes) ICA | Time (minutes) ACO-ICA |
|---|---|---|---|---|---|---|---|---|---|---|
| p01 | **576.87** | **576.87** | **576.87** | **576.87** | 586.77 | **576.87** | **576.87** | 1.3 | 4.2 | 2.5 |
| p02 | **473.53** | 475.86 | **473.53** | **473.53** | 496.46 | 475.24 | **473.53** | 1.9 | 6.2 | 3.2 |
| p03 | **641.19** | 644.46 | **641.19** | **641.19** | 672.46 | 655.29 | 644.70 | 2.6 | 7.9 | 4.2 |
| p04 | **1001.59** | 1020.52 | 1006.66 | **1001.59** | 1072.06 | 1015.11 | 1011.35 | 2.4 | 12.4 | 6.2 |
| p05 | **750.03** | 751.90 | 753.40 | 750.11 | 790.80 | 789.15 | 767.46 | 2.3 | 20.3 | 12.2 |
| p06 | **876.50** | 885.84 | **876.50** | **876.50** | 911.96 | 887.71 | 884.98 | 2.7 | 14.7 | 2.7 |
| p07 | **885.80** | 891.95 | 895.53 | 887.11 | 949.52 | 916.79 | 891.70 | 2.0 | 11.5 | 2.3 |
| p08 | **4420.94** | 4485.08 | 4482.44 | **4420.94** | 4510.81 | 4502.22 | 4470.18 | 23.7 | 60.0* | 53.6 |
| p09 | **3900.22** | 3971.59 | 3937.81 | **3900.22** | 4017.15 | 3986.70 | 3955.97 | 28.7 | 60.0* | 36.9 |
| p10 | **3663.02** | 3747.62 | 3714.65 | **3663.02** | 3806.98 | 3801.16 | 3686.46 | 33.1 | 60.0* | 43.2 |
| p11 | 3554.18 | 3599.93 | 3569.68 | **3554.08** | 3686.45 | 3644.02 | 3561.39 | 25.2 | 60.0* | 50.6 |
| p12 | **1318.95** | **1318.95** | **1318.95** | **1318.95** | 1373.52 | 1359.49 | 1360.69 | 1.9 | 10.0 | 4.6 |
| p13 | **1318.95** | **1318.95** | **1318.95** | **1318.95** | 1353.22 | 1320.79 | 1320.27 | 2.0 | 8.9 | 3.4 |
| p14 | **1360.12** | 1373.18 | 1365.68 | **1360.12** | 1419.31 | 1394.01 | 1379.70 | 2.4 | 6.7 | 4.5 |
| p15 | **2505.42** | 2588.22 | 2565.67 | **2505.42** | 2679.65 | 2644.14 | 2556.87 | 3.2 | 25.5 | 7.2 |
| p16 | **2572.23** | **2572.23** | **2572.23** | **2572.23** | 2583.22 | 2577.66 | 2579.19 | 3.2 | 16.0 | 3.4 |
| p17 | **2709.09** | 2731.37 | **2709.09** | **2709.09** | 2773.67 | 2742.93 | 2724.59 | 4.2 | 12.3 | 5.5 |
| p18 | **3702.85** | 3781.03 | 3781.03 | 3781.03 | 3922.48 | 3855.70 | 3854.31 | 29.2 | 60.0* | 40.3 |
| p19 | **3827.06** | 3831.71 | **3827.06** | **3827.06** | 3866.40 | 3857.36 | 3837.90 | 35.4 | 42.3 | 45.5 |
| p20 | **4058.07** | 4142.00 | 4097.06 | **4058.07** | 4257.20 | 4164.88 | 4168.72 | 34.5 | 60.0* | 58.0 |
| p21 | **5474.84** | 5617.53 | 5535.99 | 5495.54 | 5817.76 | 5764.61 | 5560.31 | 60.0* | 60.0* | 60.0* |
| p22 | **5702.16** | 5832.07 | 5772.23 | 5772.23 | 6047.49 | 5963.71 | 5994.10 | 60.0* | 60.0* | 60.0* |
| p23 | **6095.46** | 6183.13 | 6145.58 | 6145.58 | 6261.69 | 6295.46 | 6203.31 | 60.0* | 60.0* | 60.0* |
| Average error | | 1.20% | 0.67% | 0.20% | 4.16% | 2.59% | 1.46% | | | |

Compared with other algorithms in Table 3, ACO-ICA was able to obtain the same best score in 12 out of 23 instances and outperformed the four rival algorithms on p08 instance. On average error percentage in respect to BKS, ACO-ICA fell short compared to ACO+ (0.13% vs 0.20% error), however, outperformed other compared approaches. Therefore, it can be concluded that the proposed ACO-ICA hybrid 2-stage algorithm is appropriate for solving the MDVRP, leveraging the ICA for customer assignment and ACO for customer routing.

Table 3. Best solution obtained by ACO-ICA compared to other algorithms in the literature across Cordeau's MDVRP benchmark instances and the best-known solution (BKS). The best scores represented in bold, N representing the number of customers, M – the number of depots, $Q_{max}$ – maximum vehicle capacity, $R_{max}$ - maximum vehicle duration for a given problem instance. Average error percentage calculated using BKS as reference.

| Instance | N | M | $Q_{max}$ | $R_{max}$ | BKS [33] | CoES, 2016 [34] | IACO, 2017 [18] | TSH, 2019 [35] | ACO+, 2020 [9] | ACO-ICA (this work) |
|---|---|---|---|---|---|---|---|---|---|---|
| p01 | 50 | 4 | 80 | ∞ | **576.87** | **576.87** | **576.87** | **576.87** | **576.87** | **576.87** |
| p02 | 50 | 4 | 160 | ∞ | **473.53** | 473.87 | **473.53** | **473.53** | **473.53** | **473.53** |
| p03 | 75 | 5 | 140 | ∞ | **641.19** | **641.19** | **641.19** | **641.19** | **641.19** | **641.19** |
| p04 | 100 | 2 | 100 | ∞ | 1001.59 | 1007.40 | **1001.49** | 1008.47 | 1003.52 | 1001.59 |
| p05 | 100 | 2 | 200 | ∞ | **750.03** | 750.11 | 750.26 | 758.87 | 751.90 | 750.11 |
| p06 | 100 | 3 | 100 | ∞ | **876.50** | **876.50** | **876.50** | 881.76 | 881.60 | **876.50** |
| p07 | 100 | 4 | 100 | ∞ | 885.80 | 888.41 | 885.69 | 896.96 | **884.66** | 887.11 |
| p08 | 249 | 2 | 500 | 310 | **4420.94** | 4445.37 | 4482.44 | 4430.36 | 4428.00 | **4420.94** |
| p09 | 249 | 3 | 500 | 310 | 3900.22 | 3895.70 | 3912.23 | 3971.59 | **3897.33** | 3900.22 |
| p10 | 249 | 4 | 500 | 310 | 3663.02 | 3666.35 | 3663.00 | 3779.10 | **3657.03** | 3663.02 |
| p11 | 249 | 5 | 500 | 310 | 3554.18 | 3569.68 | 3648.95 | 3652.01 | **3549.99** | 3554.08 |
| p12 | 80 | 2 | 60 | ∞ | **1318.95** | **1318.95** | **1318.95** | **1318.95** | **1318.95** | **1318.95** |
| p13 | 80 | 2 | 60 | 200 | **1318.95** | **1318.95** | **1318.95** | **1318.95** | **1318.95** | **1318.95** |
| p14 | 80 | 2 | 60 | 180 | **1360.12** | **1360.12** | 1365.68 | 1365.69 | **1360.12** | **1360.12** |
| p15 | 160 | 4 | 60 | ∞ | **2505.42** | 2526.06 | 2505.29 | 2552.79 | **2505.42** | **2505.42** |
| p16 | 160 | 4 | 60 | 200 | **2572.23** | **2572.23** | 2587.87 | **2572.23** | **2572.23** | **2572.23** |
| p17 | 160 | 4 | 60 | 180 | 2709.09 | 2709.09 | **2708.99** | 2731.37 | 2709.09 | 2709.09 |
| p18 | 240 | 6 | 60 | ∞ | **3702.85** | 3771.35 | 3781.04 | 3802.29 | 3710.49 | 3781.03 |
| p19 | 240 | 6 | 60 | 200 | **3827.06** | **3827.06** | **3827.06** | 3831.71 | **3827.06** | **3827.06** |
| p20 | 240 | 6 | 60 | 180 | **4058.07** | **4058.07** | **4058.07** | 4097.06 | 4091.78 | **4058.07** |
| p21 | 360 | 9 | 60 | ∞ | **5474.84** | 5608.26 | **5474.84** | 5617.53 | 5505.39 | 5495.54 |
| p22 | 360 | 9 | 60 | 200 | 5702.16 | 5702.16 | **5702.06** | 5706.81 | 5702.16 | 5772.23 |
| p23 | 360 | 9 | 60 | 180 | **6095.46** | 6129.99 | **6095.46** | 6145.58 | 6140.53 | 6145.58 |
| | | | | Average error | | 0.33% | 0.33% | 0.96% | 0.13% | 0.20% |

## 5. Conclusions and future work

This work solves multi-depot vehicle routing problem (MDVRP) using novel hybrid ACO-ICA 2-stage algorithm, where Imperialist Competitive Algorithm is used for assigning customers to the depots and Ant Colony Optimization used for routing and sequencing customers across vehicles from depots. The algorithm was run across 23 common MDVRP instances and results show that proposed hybrid outperforms ACO and ICA non-hybrid counterparts. When compared to other state-of-the-art methods, ACO-ICA shows very competitive results across all benchmark instances.

However, the proposed hybrid algorithm introduces extra complexity and therefore increased computation time, which could be addressed as part of future research. Because two standalone population algorithms have been combined, the number of hyperparameters for the algorithm has also been increased, making algorithmic parameter selection harder and more time-consuming. ACO combined of with simpler algorithms, such as evolution strategy for customer grouping, could be investigated. Moreover, the current algorithm implementation assigns ACO's global pheromone matrix per ICA's country, another approach, of pheromone matrix per empire or even per population, could be investigated.

## 6. References

[1]   G. B. Dantzig and J. H. Ramser, "The Truck Dispatching Problem," *Manage. Sci.*, vol. 6, no. 1, pp. 80–91, Oct. 1959, doi: 10.1287/mnsc.6.1.80.

[2]   E. L. Lawler, J. K. Lenstra, A. H. G. Rinnooy Kan, and D. B. Shmoys, "Erratum: The Traveling Salesman Problem: A Guided Tour of Combinatorial Optimization," *J. Oper. Res. Soc.*, vol. 37, no. 6, pp. 655–655, Jun. 1986, doi: 10.1057/jors.1986.117.

[3]   S. Karakatič and V. Podgorelec, "A survey of genetic algorithms for solving multi depot vehicle routing problem," *Appl. Soft Comput. J.*, vol. 27, pp. 519–532, 2015, doi: 10.1016/j.asoc.2014.11.005.

[4]   S. Samsuddin, M. S. Othman, and L. M. Yusuf, "a Review of Single and Population-Based Metaheuristic Algorithms Solving Multi Depot Vehicle Routing Problem," *Int. J. Softw. Eng. Comput. Syst.*, vol. 4, no. 2, pp. 80–93, 2018, doi: 10.15282/ijsecs.4.2.2018.6.0050.

[5]   M. R. Garey and D. S. Johnson, *Computers and Intractability: A Guide to the Theory of NP-Completeness*, W.H. Freem. 1979.

[6]   N. Sharma and M. Monika, "A Literature Survey on Multi Depot Vehicle Routing Problem," *IJSRD -International J. Sci. Res. Dev.*, vol. 3, no. 04online, pp. 2321–613, 2015.

[7]   L. F. Galindres-Guancha, E. M. Toro-Ocampo, and R. A. Gallego-Rendón, "Multi-objective MDVRP solution considering route balance and cost using the ILS metaheuristic," *Int. J. Ind. Eng. Comput.*, vol. 9, no. 1, pp. 33–46, 2018, doi: 10.5267/j.ijiec.2017.5.002.

[8]   J. Renaud, G. Laporte, and F. F. Boctor, "A tabu search heuristic for the multi-depot vehicle routing problem," *Comput. Oper. Res.*, vol. 23, no. 3, pp. 229–235, 1996, doi: 10.1016/0305-0548(95)O0026-P.

[9]   P. Stodola, "Hybrid ant colony optimization algorithm applied to the multi-depot vehicle routing problem," *Nat. Comput.*, vol. 6, 2020, doi: 10.1007/s11047-020-09783-6.

[10]  G. Clarke and J. W. Wright, "Scheduling of Vehicles from a Central Depot to a Number of Delivery Points," *Oper. Res.*, vol. 12, no. 4, pp. 568–581, Aug. 1964, doi: 10.1287/opre.12.4.568.

[11]  P. Surekha and S. Sumathi, "Solution To Multi-Depot Vehicle Routing Problem Using Genetic Algorithms," *World Appl. Program.*, no. 13, pp. 118–131, 2011.

[12]  B. E. Gillett and J. G. Johnson, "Multi-terminal vehicle-dispatch algorithm," *Omega*, vol. 4, no. 6, pp. 711–718, 1976, doi: 10.1016/0305-0483(76)90097-9.

[13]  D. Gulczynski, B. Golden, and E. Wasil, "The multi-depot split delivery vehicle routing problem: An integer programming-based heuristic, new test problems, and computational results," *Comput. Ind. Eng.*, vol. 61, no. 3, pp. 794–804, 2011, doi: 10.1016/j.cie.2011.05.012.

[14]  A. Imran, "A Variable Neighborhood Search-Based Heuristic for the Multi-Depot Vehicle Routing Problem," *J. Tek. Ind.*, vol. 15, no. 2, pp. 95–102, Dec. 2013, doi: 10.9744/jti.15.2.95-

102.

[15] T. Dokeroglu, E. Sevinc, T. Kucukyilmaz, and A. Cosar, "A survey on new generation metaheuristic algorithms," *Comput. Ind. Eng.*, vol. 137, no. September, p. 106040, Nov. 2019, doi: 10.1016/j.cie.2019.106040.

[16] Y. M. Shen and R. M. Chen, "Optimal multi-depot location decision using particle swarm optimization," *Adv. Mech. Eng.*, vol. 9, no. 8, pp. 1–15, 2017, doi: 10.1177/1687814017717663.

[17] S. B. Sarathi Barma, J. Dutta, and A. Mukherjee, "A 2-opt guided discrete antlion optimization algorithm for multi-depot vehicle routing problem," *Decis. Mak. Appl. Manag. Eng.*, vol. 2, no. 2, pp. 112–125, 2019, doi: 10.31181/dmame1902089b.

[18] B. Yao, C. Chen, X. Song, and X. Yang, "Fresh seafood delivery routing problem using an improved ant colony optimization," *Ann. Oper. Res.*, vol. 273, no. 1–2, pp. 163–186, 2019, doi: 10.1007/s10479-017-2531-2.

[19] C.-J. Ting and C.-H. Chen, "Combination of Multiple Ant Colony System and Simulated Annealing for the Multidepot Vehicle-Routing Problem with Time Windows," *Transp. Res. Rec. J. Transp. Res. Board*, vol. 2089, no. 1, pp. 85–92, Jan. 2008, doi: 10.3141/2089-11.

[20] B. Yu, Z. Z. Yang, and J. X. Xie, "A parallel improved ant colony optimization for multi-depot vehicle routing problem," *J. Oper. Res. Soc.*, vol. 62, no. 1, pp. 183–188, 2011, doi: 10.1057/jors.2009.161.

[21] I. Dzalbs, T. Kalganova, and I. Dear, "Imperialist Competitive Algorithm with Independence and Constrained Assimilation for Solving 0-1 Multidimensional Knapsack Problem," Mar. 2020.

[22] I. Dzalbs and T. Kalganova, "Accelerating supply chains with Ant Colony Optimization across range of hardware solutions," 2020.

[23] M. Dorigo and T. Stützle, "Ant Colony Optimization: Overview and Recent Advances," in *International Series in Operations Research and Management Science*, vol. 272, 2019, pp. 311–351.

[24] E. Atashpaz-Gargari and C. Lucas, "Imperialist competitive algorithm: An algorithm for optimization inspired by imperialistic competition," *2007 IEEE Congr. Evol. Comput. CEC 2007*, pp. 4661–4667, 2007, doi: 10.1109/CEC.2007.4425083.

[25] S. Hosseini and A. Al Khaled, "A survey on the Imperialist Competitive Algorithm metaheuristic: Implementation in engineering domain and directions for future research," *Appl. Soft Comput. J.*, vol. 24, pp. 1078–1094, 2014, doi: 10.1016/j.asoc.2014.08.024.

[26] Gengjia Wang, Y. Zhang, and J. Chen, "A Novel Algorithm to Solve the Vehicle Routing Problem with Time Windows: Imperialist Competitive Algorithm," *Int. J. Adv. Inf. Sci. Serv. Sci.*, vol. 3, no. 5, pp. 108–116, Jun. 2011, doi: 10.4156/aiss.vol3.issue5.14.

[27] S. Shamshirband, M. Shojafar, A. A. R. Hosseinabadi, and A. Abraham, "OVRP_ICA: An Imperialist-Based Optimization Algorithm for the Open Vehicle Routing Problem," 2015, pp. 221–233.

[28] B. Chandra Mohan and R. Baskaran, "A survey: Ant Colony Optimization based recent research and implementation on several engineering domain," *Expert Syst. Appl.*, vol. 39, no. 4, pp. 4618–4627, Mar. 2012, doi: 10.1016/j.eswa.2011.09.076.

[29] M. Mehdinejad, B. Mohammadi-Ivatloo, R. Dadashzadeh-Bonab, and K. Zare, "Solution of

optimal reactive power dispatch of power systems using hybrid particle swarm optimization and imperialist competitive algorithms," *Int. J. Electr. Power Energy Syst.*, vol. 83, pp. 104–116, Dec. 2016, doi: 10.1016/j.ijepes.2016.03.039.

[30] S. Hosseini, A. Al Khaled, and S. Vadlamani, "Hybrid imperialist competitive algorithm, variable neighborhood search, and simulated annealing for dynamic facility layout problem," *Neural Comput. Appl.*, vol. 25, no. 7–8, pp. 1871–1885, Dec. 2014, doi: 10.1007/s00521-014-1678-x.

[31] H. Eskandar, P. Salehi, and M. H. Sabour, "Imperialist Competitive Ant Colony Algorithm for Truss Structures," *Appl. Sci.*, vol. 12, no. 33, pp. 94–105, 2011.

[32] J.-F. Cordeau, M. Gendreau, and G. Laporte, "A tabu search heuristic for periodic and multi-depot vehicle routing problems," *Networks*, vol. 30, no. 2, pp. 105–119, Sep. 1997, doi: 10.1002/(SICI)1097-0037(199709)30:2<105::AID-NET5>3.0.CO;2-G.

[33] "Multiple Depot VRP Instances," *University of Malaga, Spain.* [Online]. Available: http://neo.lcc.uma.es/vrp/vrp-instances/multiple-depot-vrp-instances/.

[34] F. B. De Oliveira, R. Enayatifar, H. J. Sadaei, F. G. Guimarães, and J. Y. Potvin, "A cooperative coevolutionary algorithm for the Multi-Depot Vehicle Routing Problem," *Expert Syst. Appl.*, vol. 43, pp. 117–130, 2016, doi: 10.1016/j.eswa.2015.08.030.

[35] M. E. H. Sadati, D. Aksen, and N. Aras, "The r-interdiction selective multi-depot vehicle routing problem," *Int. Trans. Oper. Res.*, vol. 27, no. 2, pp. 835–866, 2020, doi: 10.1111/itor.12669.